\documentclass[10pt, final, conference, letterpaper, twoside, twocolumn]{IEEEtran}
\usepackage{soul}
\usepackage{multirow}
\usepackage{booktabs} 
\usepackage{ragged2e}
\usepackage[noend]{algpseudocode}     
\usepackage{algorithm}            
\usepackage{threeparttable}
\usepackage{comment}
\usepackage{graphicx}
\usepackage{amsmath}
\usepackage{cite}
\usepackage{enumitem}
\usepackage[T1]{fontenc}
\usepackage[utf8]{inputenc}

\usepackage[table,dvipsnames,svgnames]{xcolor}
\usepackage{colortbl}
\usepackage[caption=false,font=normalsize,labelfont=sf,textfont=sf]{subfig}

\usepackage[hyphens]{url}
\usepackage{breakurl}
\usepackage[hidelinks]{hyperref}
\usepackage{xspace}

\algrenewcommand\algorithmicforall{\textbf{foreach}}

\usepackage{setspace}
\definecolor{mygreen}{RGB}{0,128,0} 

\newcommand{\extended}[1]{{\color{mygreen}{#1}}}
\renewcommand{\extended}[1]{#1}

\newcommand{\jk}[1]{{\color{blue}{#1}}}

\newcommand{\jkc}[1]{{\color{blue}JK: {#1}}}
\renewcommand{\jkc}[1]{}

\newcommand{\la}[1]{{\color{blue}LA: {#1}}}
\renewcommand{\la}[1]{}

\newcommand{\zw}[1]{{\color{mygreen}{#1}}}
\renewcommand{\zw}[1]{#1}

\newcommand{\zwnew}[1]{{\color{mygreen}{#1}}}
\renewcommand{\zwnew}[1]{#1}

\usepackage[most]{tcolorbox}
\newtcolorbox{mybox}[2][]{
  title=#2,
  coltitle=black,             
  colbacktitle=gray!30,       
  colback=white,              
  colframe=black,             
  arc=5pt,                    
  outer arc=5pt,              
  boxrule=0.5pt,              
  enhanced,                   
  attach boxed title to top left={yshift=-2.5mm, xshift=2mm}, 
  boxed title style={
    size=small,
    colback=gray!30,         
    colframe=black,          
  },
  drop shadow=black!50!white, 
  top=1.8mm,                    
  bottom=0.2mm,                 
  left=1mm,                   
  right=1mm,                  
  #1
}

\begin{document}

\newcommand{\thetitle}{LLMs and the Future of Chip Design:\\
	Unveiling Security Risks and Building Trust}

\title{%
\vspace{-0.55em}
\thetitle}

\author{%
Zeng~Wang,$^\dag$
Lilas~Alrahis,$^\ddag$ Likhitha~Mankali,$^\dag$ Johann~Knechtel,$^\ddag$~and~Ozgur~Sinanoglu~$^\ddag$
\\
\IEEEauthorblockA{
$^\dag$New York University, USA\\
$^\ddag$New York University Abu Dhabi, UAE\\
\normalsize{Email:\{zw3464, lma387, likhitha.mankali, johann, ozgursin\}@nyu.edu}}%
\vspace{-2em}}

\IEEEtitleabstractindextext{
\begin{abstract}
Chip design is about to be revolutionized by the integration of large language, multimodal, and circuit models (
collectively LxMs). 
While exploring this exciting frontier with
	tremendous potential, the community must also carefully consider the related security risks and the need for building trust into using LxMs for chip design.
	First, we review the recent surge of using LxMs for chip design in general. We cover state-of-the-art works for the automation of hardware description language code generation and for scripting and guidance of
	essential but cumbersome tasks for electronic design automation tools, e.g., design-space exploration, tuning, or designer training.
	Second,
we raise and provide initial answers to novel research questions
	on critical issues for security and trustworthiness of LxM-powered chip design from both the attack and defense perspectives.
\end{abstract}

}

\maketitle
\IEEEdisplaynontitleabstractindextext
\IEEEpeerreviewmaketitle

\section{Introduction}
\label{sec:intro}

The application of artificial intelligence (AI) in chip design workflows, particularly within electronic design automation (EDA), has significantly accelerated the design process in recent years for critical tasks such as
high-level synthesis (HLS), etc.~\cite{rapp2021mlcad}.
Besides, recent advancements in natural language processing (NLP) and large language models (LLMs) have demonstrated remarkable success in tasks such as creative writing, chatbot interaction, and commonsense reasoning~\cite{openai2023gpt, touvron2023llama}. Given the similarities for NLP applications and certain tasks in EDA, including pattern recognition, code generation, and more, along with the demonstrated success of other AI models in EDA, researchers have recently begun to investigate the potential of LLMs across various aspects of EDA.\footnote{%
In fact, LLMs are already used for commercial EDA applications; e.g.,
\textit{Synopsys.ai Copilot} is a generative AI tool that aids in reducing time to market and systemic complexity through conversational intelligence~\cite{synopsys}.}

In this paper, we explore a broad spectrum of critical issues, presenting selected research that focuses on the use of LLMs, large multimodal models (LMMs), and large circuit models (LCMs), or collectively LxMs, for chip design, verification, and security analysis. Our contributions are as follows.

\begin{enumerate}[leftmargin=*]
 \item We provide a comprehensive survey, reviewing selected state-of-the-art and providing detailed descriptions of LLM tasks, framework designs, training scheme, etc.
 \item We compile resources on LLMs for chip design, focusing on various applications of LLMs, security concerns, and open sources. We summarize these findings in our \textbf{\textit{LLM4IC} hub} [\url{https://github.com/DfX-NYUAD/LLM4IC}].
\end{enumerate}

\section{Background}

\subsection{Large Language Models}

LLMs are transformer-based deep neural networks. The transformer architecture comprises encoder and decoder layers that utilize self-attention mechanisms to process input sequences, such as text sentences or paragraphs. These models are trained on massive text datasets and are capable of performing various NLP tasks, including text recognition, translation, prediction, and generation. 
\subsubsection{Learning Paradigms}  We focus on four learning/usage paradigms in the field of LLMs, as follows.

\textbf{Training.} Generic LLMs are trained on large-scale datasets and can be directly used for various tasks without modification~\cite{openai2023gpt}. \textit{Domain-adaptive pretraining} (DAPT) involves pretraining the LLM on domain-specific text, allowing it to fully adapt to the specifics of a particular domain~\cite{liu2023chipnemo}.

\textbf{Prompting} involves crafting specific prompts that guide the pretrained LLM in generating desired outputs without modifying the model. 
Prompts that guide LLM with a few examples or illustrations are known as \textit{few-shot}. Similarly, \textit{one-shot} uses one example, and \textit{zero-shot} uses none~\cite{wei2021finetuned}.

\textbf{Fine-tuning (FT)} is the process of further training a pretrained LLM on \zwnew{task-specific} data to improve its performance on \zwnew{these seen tasks}~\cite{openai2023gpt}. 
{Note that this differs from DAPT.}

\textbf{Instruct-tuning (IT)} is the process of enhancing the performance of pretrained LLMs by further training them \zwnew{with data for instruction-and-example pairs to handle unseen tasks~\cite{zhang2023instruction}.}

\extended{\subsubsection{Interacting Mechanisms}
Retrieval-augmented generation (RAG) is a mechanism that extracts queries from input prompts and then fetches relevant
	information from external databases. This information is used by a generative component/agent to produce follow-up responses to the LLM~\cite{lewis2020retrieval}.
		Besides, the reasoning and action (ReAct) mechanism
			prompts LLMs to
		provide both reasoning traces and actions related to a task in an interleaved fashion, thereby stimulating the model's dynamic problem-solving abilities~\cite{yao2022react}.
}

\subsubsection{LMM/LCM} 
LMMs integrate multiple data types (e.g., text, images) and various neural network architectures.
This enables them to effectively handle diverse data with complex relations often found in real-world problems~\cite{wu2023multimodal}. Similarly, LCMs integrate multiple hardware representations of chip designs to
work on
complex chips and electronic systems~\cite{chen2024dawn}.

\subsection{Electronic Design Automation}

EDA utilizes a range of specialized software tools to realize automated work-flows for chip design.
Detailed technology information as well as
reusable design components, so-called intellectual property (IP) modules, are utilized as well.
A typical EDA flow incorporates distinct stages as follows.\footnote{%
Contemporary flows also include verification procedures as well as feedback loops across each stage; for simplicity, this
is not further discussed here.
Besides, see also~\cite{KLMH11} for more background on EDA in general and~\cite{knechtel20_Sec_EDA_DATE} for a security-centric review of EDA.}

First, a system specification
is translated or compiled by a HLS tool into a hardware description language (HDL), typically at the abstraction of register-transfer
level (RTL) instructions. Toward that end, HLS tools initially devise and/or allocate functional units, both possibly with the help of IP modules. Then, HLS tools bind all the system-level task to those units. In other words, the chip's operation is planned out into a behavioural model described in HDL/RTL.

Second, that behavioural HDL/RTL model is translated by a logic-synthesis tool into a gate-level netlist. During this process, technology mapping is applied, \textit{i.e.}, the generic design description is translated into a technology-specific
one.

Third, the netlist is translated by a physical-synthesis tool into an actual layout.
This process is
known to be
the most challenging task for chip design, especially for advanced technology nodes with complex
geometries and design rules.

\subsection{Hardware Security}

The increase in chip manufacturing costs due to continued advances in technology nodes has led design companies to adopt a globalized supply chain. In such a so-called fabless business model, design companies outsource fabrication, testing, and packaging to third-party facilities. However, outsourcing to potentially untrusted entities has led to different security threats. These threats could be enacted either at different stages of the supply chain or by end-users~\cite{rostami2014primer, knechtel20_Sec_Emerg_ISPD}.

\subsubsection{Hardware Trojans (HTs)} Attackers located either in the fabrication facility (foundry) or with a third-party design house could insert malicious modifications into the design.

\subsubsection{IP Piracy and Chip Overbuilding} Attackers located in the foundry may illegally pirate the design IP or overproduce chips without the consent of the designer.

\subsubsection{Side-Channel Attacks (SCA)} Attackers can extract secret information by exploiting physical information from chips in operation, such as
power consumption, timing behavior, etc.

\subsubsection{Further Considerations}
There are other inherent vulnerabilities (\textit{i.e.}, flaws in the design) that can result in security breaks. For instance, vulnerabilities in micro-architecture design have raised severe implications~\cite{spectre,meltdown}.

Besides, any hardware vulnerabilities discovered after fabrication usually cannot be fixed/patched.
Thus, researchers have proposed different vulnerability detection techniques, such as fuzzing~\cite{fuzzing}, information flow tracking~\cite{ift}, and formal verification~\cite{formal-verification}. These techniques often utilize assertions to detect hardware vulnerabilities~\cite{assertion-check}.

\section{Proposed Taxonomy}

\begin{figure}[!t]
 \centering
 \vspace{-0.5em}
 \includegraphics[width=0.95\columnwidth]{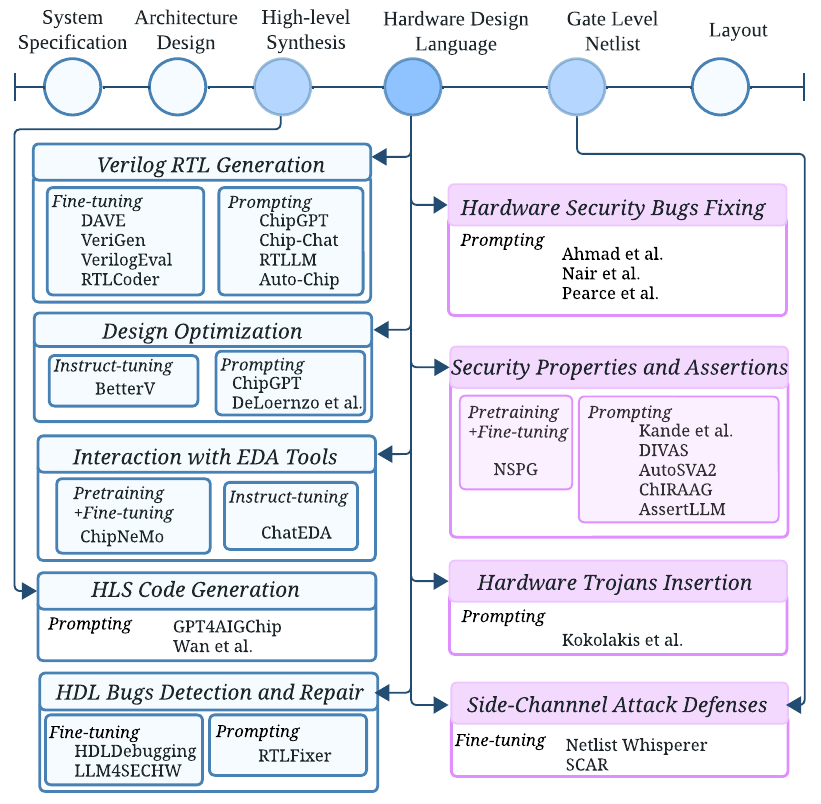}
 \caption{Overview of selected LLM applications for EDA in general (blue) and for hardware security in particular (purple).}
 \label{fig:LLM_EDA}
\end{figure}

 In this section, we discuss selected works on LLMs for various chip-related tasks, categorizing them primarily into two parts, as shown in Fig.~\ref{fig:LLM_EDA}: LLMs for EDA (Section~\ref{sec:llmeda}) and LLMs for hardware security (Section~\ref{sec:LLMSecu}). We further summarize the key details of the discussed works in Table~\ref{tab:my-table}.

\subsection{LLMs for Electronic Design Automation}
\label{sec:llmeda}

LLMs are recently explored to support chip design, focusing primarily on HLS and HDL code generation. 
Still, researchers also started to investigate
LLMs for other tasks, like generating architecture specifications~\cite{li2024specllm}, HDL bug fixes~\cite{tsai2023rtlfixer}, streamlining the printed circuit board design~\cite{han2023new}, etc.

Here, we review prior art that applies LLMs for a variety of such tasks, and we also outline related limitations.

\begin{table*}[!tb]
\vspace{-2.5em}
\caption{Summary of LLMs for Chip Design and Hardware Security.\\
PT: pretraining, P: prompting, FT: fine-tuning, IT: instruct-tuning, NA: not applicable.}
\label{tab:my-table}
\centering
\footnotesize
\setlength\tabcolsep{4pt} 
\renewcommand\arraystretch{0.66}

\begin{tabular}{llm{2.55cm}m{1cm}>{\raggedright\arraybackslash}m{2.98cm}>{\raggedright\arraybackslash}m{4.2cm}m{1cm}}
\hline

\textbf{Platform} & \textbf{LLM Task} & \textbf{Base Model} & \textbf{Training} & \textbf{FT/IT Datasets} & \textbf{Expertise Needed} &\textbf{Access} \\ \hline 

\textbf{ChipNeMo~\cite{liu2023chipnemo}} &
& LLaMA2-70B & PT+ FT & 24.1B samples from Nvidia's internal database &For training dataset generation and evaluation& Closed \\ 

\rowcolor{Gainsboro}
\textbf{ChatEDA~\cite{wu2024chateda}} &
\cellcolor{WhiteSmoke}\multirow{-4}{*}{\ \begin{tabular}[c]{@{}l@{}}\cellcolor{white}Interaction\\\cellcolor{white}with EDA Tools\end{tabular}}
& LLaMA2-70B & IT & $\approx$1,500 instructions &For EDA expert knowledge& Closed \\ \hline 

\textbf{DAVE~\cite{pearce2020dave}} &
& GPT-2 & FT & > 17K customized task-and-result pairs &For training dataset generation & Closed
\\ 

\rowcolor{Gainsboro}
\textbf{VeriGen~\cite{thakur2023verigen}} &\cellcolor{WhiteSmoke} & CodeGen-16B & FT & 400MB Verilog from GitHub \& Textbooks  & For evaluation and hand-designed prompts& \cite{vgencode} \\ 

\textbf{VerilogEval~\cite{liu2023verilogeval}} & & VeriGen & FT &  8,502
synthetic problem-and-code pairs & For training dataset generation  & \cite{vevalcode} \\ 

\rowcolor{Gainsboro}
\textbf{RTLCoder~\cite{liu2023rtlcoder}} &\cellcolor{WhiteSmoke} & Mistral-7B-v0.1, DeepSeek-Coder-6.7b& FT & 27K instruction-and-code pairs &For keywords preparation in training dataset& \cite{rtlcodercode} \\ 

\textbf{Chip-Chat~\cite{blocklove2023chip}} & & GPT-4 & P & NA & For syntax check and verification& \cite{chipchatcode} \\ 

\rowcolor{Gainsboro}
\textbf{AutoChip~\cite{thakur2023autochip}} &\cellcolor{WhiteSmoke} & GPT-3.5, GPT-4 & P & NA &To compile files, for simulation& \cite{autochipcode}\\ 

\textbf{RTLLM~\cite{lu2024rtllm}} &
\multirow{-15}{*}{\ \begin{tabular}[c]{@{}l@{}}\cellcolor{white}Verilog RTL\\\cellcolor{white}Generation\end{tabular}}
& GPT-4 & P & NA &For syntax checks, and for functionality and quality evaluation & \cite{rtllmcode}\\ \hline 

\rowcolor{Gainsboro}
\textbf{ChipGPT~\cite{chang2023chipgpt}} &\cellcolor{WhiteSmoke} 
& GPT-3.5 & P & NA & For human feedback with manual design correction& Closed \\ 

\textbf{BetterV~\cite{pei2024betterv}} & & CodeLlama-7B-Instruct & IT & a set of Verilog-and-C pairs and a set of Verilog definition-and-body pairs & For training dataset preparation and generative discriminator & Closed
\\ 

\rowcolor{Gainsboro}
\textbf{DeLorenzo et al.~\cite{delorenzo2024make}} &\cellcolor{WhiteSmoke} 
\multirow{-7}{*}{\ \begin{tabular}[c]{@{}l@{}}\cellcolor{white}Design\\\cellcolor{white}Optimization\end{tabular}}
& VeriGen-2B & P & NA &For optim.\ and manual prompts& Closed \\ \hline 

\textbf{GPT4AIGChip~\cite{fu2023gpt4aigchip}} & 
& GPT-4 & P & NA &For demo library construction and final manual design corrections&
Closed \\ 

\rowcolor{Gainsboro}
\textbf{Wan et al.~\cite{wan2024software}} &\cellcolor{WhiteSmoke} 
\multirow{-4}{*}{\ \begin{tabular}[c]{@{}l@{}}\cellcolor{white}HDL Code\\\cellcolor{white}Generation\end{tabular}}
& GPT-4 & P & NA &For setting bugs and insertion rules & ~\cite{hdlbugcode} \\ \hline 

\textbf{RTLFixer~\cite{tsai2023rtlfixer}} &
& GPT-3.5 & P & NA &For retrieval database (syntax check) & Closed \\ 

\rowcolor{Gainsboro}
\textbf{HDLdebugger~\cite{yao2024hdldebugger}} &\cellcolor{WhiteSmoke} & CodeLlama-13b & FT & 92,143 distinct Huawei's HDL instances  &For dataset generation and doc-and-code RAG & Closed \\ 

\textbf{LLM4SecHW~\cite{fu2023llm4sechw}} &
\multirow{-5}{*}{\ \begin{tabular}[c]{@{}l@{}}\cellcolor{white}HDL Bugs\\\cellcolor{white}Detection\\\cellcolor{white}and Repair\end{tabular}}
& Falcon-7B, StableLM-7B, LLaMA2-7B & FT & 15,000 samples of commit message and pre/post-revision code &For hardware version control training dataset& \cite{llm4seccode} \\ \hline 

\rowcolor{Gainsboro}
\textbf{Pearce et al.~\cite{pearce2023examining}} &\cellcolor{WhiteSmoke} 
& GPT-2 & P & NA &For bug local.\ \& fixing suggestions & \cite{zeroexamcode}\\ 

\textbf{Nair et al.~\cite{nair2023generating}} & & GPT-3.5 &P & NA & For manual security-driven prompts & Closed \\ 

\rowcolor{Gainsboro}
\textbf{Ahmad et al.~\cite{ahmad2024hardware}} &\cellcolor{WhiteSmoke} 
\multirow{-4}{*}{\ \begin{tabular}[c]{@{}l@{}}\cellcolor{white}Security\\\cellcolor{white}Bugs Fixing\end{tabular}}
& code-davinci-002 & P & NA &For bug local.\ and repair instruction & \cite{Ahmadcode} \\ \hline 

\textbf{NSPG~\cite{meng2023unlocking}} &
& BERT & PT+ FT &4,427 sentences from hardware
documentations&For labeled security property dataset& Closed \\ 

\rowcolor{Gainsboro}
\textbf{DIVAS~\cite{paria2023divas}} &\cellcolor{WhiteSmoke} & GPT-4, BARD & P & NA &For filtering related CWEs& Closed \\ 

\textbf{AutoSVA2~\cite{orenes2023rtl}} & & GPT-4 & P & NA &For SVA safety rules refinement& \cite{autosva2code} \\ 

\rowcolor{Gainsboro}
\textbf{Kande et al.~\cite{kande2024security}} &\cellcolor{WhiteSmoke} & code-davinci-002 & P & NA &For selecting designs with CWEs& Closed \\ 

\textbf{ChIRAAG~\cite{mali2024chiraag}} & & GPT-4 & P & NA &For manual error checks& Closed\\ 

\rowcolor{Gainsboro}
\textbf{AssertLLM~\cite{fang2024assertllm}} &\cellcolor{WhiteSmoke} 
\multirow{-9}{*}{\ \begin{tabular}[c]{@{}l@{}} \cellcolor{white}Assertions\\\cellcolor{white}and Security\\\cellcolor{white}Properties\\\cellcolor{white}Generation \end{tabular}}
& GPT-4 & P & NA &For SVA-related data extraction& \cite{assertllmcode} \\ \hline 

\textbf{Kokolakis et al.~\cite{kokolakis2024harnessing}} &\ \begin{tabular}[c]{@{}l@{}}HT Insertion\end{tabular} & ChatGPT & P & NA & For manual designed prompts& Closed \\ \hline 

\rowcolor{Gainsboro}
\textbf{Netlist Whisperer~\cite{nair2023netlist}} &\cellcolor{WhiteSmoke} 
& GTP-3 (1x Ada-based, 1x Curie-based) & FT & 10K secure-vs-leaking tokens, 3.5k~algebraic-normal-form pairs&For training dataset generation& Closed \\ 

\textbf{SCAR~\cite{srivastava2024scar}} &
\multirow{-5}{*}{\ \begin{tabular}[c]{@{}l@{}}\cellcolor{white}SCA\\\cellcolor{white}Counter-\\\cellcolor{white}measures\end{tabular}}
& Falcon-7B & FT & synthetic vulnerable lines from AES HDL codes&For leakage detection, local.& Closed \\ \hline 

\end{tabular}%
\vspace{-1.5em}
\end{table*}

\subsubsection{Verilog RTL Generation}

The promising code generation ability of LLMs led researchers to investigate the potential for Verilog RTL generation, aiming to relieve the burden of manual design efforts. However, there are challenges due to the limited hardware-related knowledge with existing pretrained LLMs. 
To address these challenges, researchers proposed various prompting-based and FT-based frameworks.\footnote{%
Recall that prompting-based frameworks iteratively guide pretrained models with external knowledge to deal with
limited training datasets and avoid further training, whereas FT-based frameworks address the challenge directly and
build training datasets to
fine-tune pretrained LLMs.}

\textbf{1a)~Prompting-based Frameworks}: Chip-Chat~\cite{blocklove2023chip} interacts with conversational generative pretrained transformer (GPT)-4 with prompts, alongside experienced engineers, to collaboratively design an 8-bit accumulator-based microprocessor. \zw{This work shows the possibility of going from design specifications to a functional implementation with minimal effort, but it lacks verification and modern security features.}
 
ChipGPT~\cite{chang2023chipgpt} \zw{interacts with GPT-3.5 through prompts and finalizes a design considering its functional correctness and Power/Performance/Area (PPA) optimization.}
\extended{This method is tested on simple designs, 
demonstrating improvements in area and correctness of the generated simple design.}

Further enhancing the quality of generated designs, RTLLM~\cite{lu2024rtllm} \zw{covers three objectives: syntax, functionality, and design quality.
It evaluates the capability of GPT-3.5 and GPT-4 in generating
RTL codes for these objectives and then improves performance by requiring LLMs to provide reasoning steps and syntax error checking in initial prompts.}

AutoChip~\cite{thakur2023autochip} utilizes a feedback loop to integrate with generated Verilog compilation and simulation results for more accurate code generation without human intervention. \extended{Although the
	iterative feedback improves the success rate, \zw{it cannot handle larger designs due to prompting limitations.}}

\textbf{1b)~Fine-tuning-based Frameworks}: 
DAVE~\cite{pearce2020dave} fine-tunes GPT-2 \zw{with a dataset of English-task-and-Verilog-result pairs} for deriving Verilog snippets from English. 

VeriGen~\cite{thakur2023verigen} fine-tunes the CodeGen-16B~\cite{nijkamp2022codegen} LLM using compiled Verilog datasets sourced from GitHub and textbooks in an unsupervised manner.
\zw{Its generated Verilog codes still contain functional errors, requiring designers to debug and refine them.} \zw{For instance},
VeriGen
on \zw{complex} RTLLM~\cite{lu2024rtllm} benchmarks underperforms compared to GPT-3.5~\cite{liu2023rtlcoder}.

\zw{VerilogEval~\cite{liu2023verilogeval} further fine-tunes VeriGen
with synthetic description-and-code pairs, demonstrating improved performance.
It open-sources two synthetic benchmark datasets, called VerilogEval-machine and VerilogEval-human. However, it ignores downstream task performance.}

RTLCoder~\cite{liu2023rtlcoder} provides various \zwnew{instruction-and-Verilog pairs} to fine-tune a LLM based on code quality feedback.
\extended{It utilizes quantized parameters with minimal loss of accuracy, allowing it to serve as a local assistant for engineers.}
\zw{However, it underperforms compared to GPT-4 on RTLLM~\cite{lu2024rtllm} and encounters functionality errors on complex designs.}

\subsubsection{Design Optimization}
LLMs for Verilog design optimization are also being investigated. ChipGPT~\cite{chang2023chipgpt} proposes a post-LLM search to find generated designs with optimal PPA.

BetterV~\cite{pei2024betterv} utilizes \zwnew{domain-specific} instruct-tuning methods to fine-tune LLMs\zwnew{ for Verilog domain knowledge. Its generative discriminator, trained with augmented data from fine-tuned LLMs, serves for optimized Verilog implementations.}

DeLorenzo et al.~\cite{delorenzo2024make} integrate Monte Carlo tree search to guide VeriGen~\cite{thakur2023verigen} in producing optimized designs for adders, multipliers and multiply-accumulate units of different sizes. 

\subsubsection{HLS Code Generation}
GPT4AIGChip~\cite{fu2023gpt4aigchip} \zw{employs decoupled HLS design templates and demo-augmented prompts for GPT-4 to generate AI accelerator designs. They then test the accelerator efficiency on six different neural networks. However, the need for demo prompts and the human involvement for verification limits the effectiveness.}

\zw{Wan et al.~\cite{wan2024software} use GPT-4 to insert bugs into open-source HLS designs utilizing refined prompts. However, theirs is limited by predefined bugs and \zw{response structures}.}

\subsubsection{Interaction with EDA Tools}

ChatEDA~\cite{wu2024chateda} \zwnew{applies instruct-tuning to train its model called AutoMage, which then decomposes complex EDA tasks into sub-tasks and generates corresponding EDA scripts for RTL-to-layout flows.}
\zw{While those scripts can be executed directly, their performance hinges on the specific tools setup, and generalization is elusive.}

ChipNeMo~\cite{liu2023chipnemo} employs \zw{DAPT} techniques to improve LLM performance in three specific applications: engineering assistant chatbot, EDA tool script generation, and bug summarization and analysis.
It integrates RAG to minimize errors and utilizes domain-specific fine-tuning to enhance accuracy.
\extended{The resulting DAPT models are shown to surpasses standard LLMs in domain-specific benchmarks and expert assessments.}

\subsubsection{Detection and Repair of HDL Bugs}

RTLFixer~\cite{tsai2023rtlfixer} introduces a debugging framework that combines the RAG and ReAct prompting mechanisms, enabling LLMs to function as agents for interactively debugging code with feedback.
\zw{The iterative code refinement succeeds to fix simple implementation errors but still lacks advanced reasoning and problem-solving skills for complex problems.}

HDLDebugger~\cite{yao2024hdldebugger} identifies that RTLFixer
falls short in meeting industry standards.
\zw{Thus, HDLDebugger introduces a fine-tuning process to enhance debugging capabilities.}
\zw{It supervises fine-tuning of the LLM through optimized HDL documents and RAG from search engines, along with generated thoughts for buggy code. However, handling complex scenarios is still challenging due to the limited scope of LLMs and the high cost of data gathering.}

LLM4SECHW~\cite{fu2023llm4sechw} \zw{fine-tunes LLMs on a debugging-oriented dataset, compiled from version control data of open-source designs and their remediation steps. This dataset address the scarcity of functional hardware designs and enables fine-tuning of LLMs for detecting and repairing hardware bugs. Its subtle
performance indicates the need for
an even larger dataset to improve effectiveness.}

\subsection{LLMs for Hardware Security}
\label{sec:LLMSecu}

Here, we discuss selected applications of LLMs for hardware security, including bug fixes, security properties and assertions generation, and handling threats like HTs and SCAs.
As before, we outline related limitations as well.

\subsubsection{Fixing of Hardware-Security Bugs}
\zw{Pearce et al.~\cite{pearce2023examining} explore LLMs for repairing vulnerabilities in a zero-shot setting. This work showcases that the considered black-box LLMs can indeed generate fixes without additional training, albeit they require strategically devised prompts.}
Also, theirs is ineffective for generating plausible fixes related to real-world scenarios and, thus, is not reliable yet for critical fixes.

Nair et al.~\cite{nair2023generating} systematically explore strategies to guide ChatGPT for secure HDL generation \zw{under 10 well-known CWEs~\cite{mitre2021cwe1194}}. This work first shows the potential security vulnerabilities in the code generated
and then proposes techniques for creating robust prompts that generate secure codes.

Ahmad et al.~\cite{ahmad2024hardware} explore the use of LLMs for automatically repairing security bugs. They evaluate repairs based on functionality and security gains.
They \zw{utilizes domain knowledge of CWE bugs, along with proposed fixing instructions, to generate prompts.
The limited use of generative instructions and the lack of a comprehensive functionality-versus-security evaluation, however, impacts the performance of such LLM-based fixes.}
\zw{Besides, this work shows that fixes which span multiple lines or require removing buggy lines are elusive.}

\subsubsection{Generation of Security Properties and Assertions}

NSPG~\cite{meng2023unlocking} is an NLP-based security property generator,
\zw{extracting properties from system-on-chip (SoC) documentation.}
\zw{It pretrains the BERT~\cite{devlin2018bert} model with SoC documentations and then fine-tunes the model with customized hardware-domain datasets. Thus, its effectiveness relies on the quality of the documentation and the scope of the datasets.}

Kande et al.~\cite{kande2024security} evaluate LLMs for security assertion generation. \zw{They generate prompts using designed templates, to guide LLMs to produce assertions, which are
then used in a SystemVerilog simulator. The simulation logs are parsed for assertion violations and analyzed against a set of ``golden assertions.''}
\zw{Designer expertise is needed to analyze the security properties and to accurately assess their effectiveness.}

DIVAS~\cite{paria2023divas}
\zw{leverages LLMs to identify relevant CWEs~\cite{mitre2021cwe1194} within selected SoC specifications, then generates SystemVerilog assertions (SVAs) using LLMs for verification, all toward automated, policy-based SoC design flows. While identifying CWEs using LLMs is feasible, this works also shows that there are limitations for accuracy and scope.}

AutoSVA2~\cite{orenes2023rtl} combines formal verification and GPT-4 to generate SVAs directly from RTL, \textit{i.e.,} without relying on specifications. This method successfully detects a previously unknown bug in a RISC-V Ariane SoC. \zw{It shows the potential to fine-tune LLMs using curated datasets of SVAs and RTL.}

ChIRAAG~\cite{mali2024chiraag} systematically deconstructs design specifications into standardized texts and then uses LLMs to generate assertions from those. It also generates testbenches to validate the generated assertions, \extended{serving as feedback to enhance the LLM's performance}.
\zw{Still, further exploration is needed to assess both the quality of unstructured specifications and the extent of functional coverage of assertions.}

AssertLLM~\cite{fang2024assertllm} automatically generates assertions.
It simplifies the process into three steps: extracting relevant details from the specifications, aligning signal names with their definitions in HDL codes, and generating SVAs. This approach utilizes three specialized LLMs to handle each step separately.
\zw{Naturally, the quality of the input specification and the effectiveness of the LLMs dictate the quality of the generated SVAs.}

\subsubsection{HT Insertion}
Kokolakis et al.~\cite{kokolakis2024harnessing} study the use of LLMs to assist attackers for HT insertion into SoC designs. Their method starts with a filtering process \zw{to identify suitable modules for HT insertions.} 
Next, the attacker provides RTL codes of the selected modules into the LLM, which then aids for implanting the HTs by modifying the modules' codes.
\zw{The approach for automated insertion with fine-tuned LLMs still requires further exploration to support complex scenarios.}

\subsubsection{SCA Countermeasures}

Netlist Whisperer~\cite{nair2023netlist} utilizes LLMs to identify SCA vulnerabilities in Verilog netlists based on a two-phase, pre-silicon flow. First, a \zw{fine-tuned} Ada-based GPT-3 model \zw{identifies problematic nets that induce power leakage.}
Second, \zw{a fine-tuned} Curie-based GPT-3 model generates a SCA-resistant version of the netlist. This approach streamlines and speeds-up the SCA assemment and fixing process, by eliminating the need for power trace collection.

SCAR~\cite{srivastava2024scar} converts the RTL codes of cryptography accelerators to control-data flow graphs, to identify modules susceptible to SCA leakage, and employs a deep-learning explainer for detection and localization of vulnerabilities. It then utilizes a pretrained LLM to automatically generate and insert additional code into the localized areas.

\section{Discussion and Future Directions}
Here, we raise important and novel research questions. We also outline possible answers/solutions and future directions.

\subsection{Research Questions}
\begin{mybox}{Research Question 1}
Can malicious modifications (HTs) be more easily introduced by integration of LLMs into design flows?
\end{mybox}
HTs can indeed be introduced through prompting by attackers~\cite{kokolakis2024harnessing}, albeit with manual efforts.
Still, leveraging the multi-reasoning capabilities
of LLMs, it seems feasible to automate.
{This complex task could be sub-divided, e.g., locating vulnerable design parts, inserting HTs, verifying their functional correctness, and testing their effectiveness.
Prompts can be augmented with related security assessment and/or simulation results toward more successful HT insertion.}

\begin{mybox}{Research Question 2}
Can valuable design IP, from sensitive in-house training data, be leaked through LLMs?
\end{mybox}
Fine-tuning pretrained LLMs with domain-specific data enables more effective RTL codes~\cite{liu2023rtlcoder,pei2024betterv}. Thus, to achieve high-quality RTL generation, vendors may have to fine-tune these models using valuable IP designs. However, this process could inadvertently expose the actual IP to attackers that seek to extract the training data.
In fact, this privacy risk has already been studied by the machine learning community, e.g., \cite{pan2020privacy} showed that an adversary with no prior knowledge can achieve $\approx$75\% accuracy when retrieving sensitive medical data from BERT responses.
\zw{Privacy preserving mapping~\cite{salamatian2015managing} can be applied to counter such adversaries; such approach should also be applicable to protect sensitive IP for EDA-centric LLMs.}

\begin{mybox}{Research Question 3}
Can LLMs be trained to seamlessly integrate best practices for hardware security into the design flow, minimizing human error and vulnerabilities against various attacks?
\end{mybox}
The success of HDL debugging with LLMs~\cite{yao2024hdldebugger}, facilitated by synthetic supervised fine-tuning, highlights the potential for similar work in enhancing hardware security in general. Designers could create datasets with pairs of attacks and suitable countermeasures. By fine-tuning LLMs with such specialized datasets, the models could become adept at not only recognizing security vulnerabilities but also generating secure designs. Besides, Netlist Whisperer~\cite{nair2023netlist} presents an alternative strategy in which LLMs are fine-tuned to pinpoint vulnerabilities and, once identified, another fine-tuned LLM could help to fix those to generate more secure designs.

\begin{mybox}{Research Question 4}
Can LLMs assist in devising better design-for-trust techniques such as advanced root-of-trust modules or novel logic-locking schemes?
\end{mybox}
LLMs have shown potential in identifying the security properties of chip designs~\cite{meng2023unlocking}. This opens up the possibility of fine-tuning LLMs to discover previously unrecognized properties of secure designs. Designers may not fully understand why a particular design is sufficiently secure---the properties identified by LLMs can provide valuable insights, helping designers harden their chips to the fullest extent.
\subsection{Future Directions}
The current application of LLMs has demonstrated their formidable capabilities for chip design in general as well as for hardware security in particular. However, numerous aspects require further exploration to fully integrate LxMs--not only LLMs--into the chip design domain.

Below, we outline several key research directions:
\subsubsection{Representation Alignment}\zw{Recently, LMMs and LCMs offer to integrate diverse formats from various design stages, such as specifications, HLS, RTL, etc., to facilitate a more accurate and coherent foundation for AI-assistance with complex design tasks, like verification, innovative design-space exploration, etc.~\cite{wu2023multimodal,chen2024dawn}.}
We call for more works utilizing LxMs toward such highly promising alignment for chip design in general and hardware security in particular.
\subsubsection{Optimization} \zw{Techniques for enhancing LxM performance remain underexplored, particularly in balancing security and functionality. For example, generative design can be approached as a search space primarily concerned with functional correctness. Then, selecting and combining generated modules from such space should also account for security.}
\subsubsection{Design Automation} HDL code generation and verification is still underdeveloped. \zw{For example, to generate complex, yet correct designs, and to identify possible CWEs, all without designer guidance and expertise, has yet to be achieved.}
\subsubsection{LxM Security} Threats to LxMs themselves, including the sensitive nature of datasets and fine-tuned models, requires thorough efforts. For example, using untrusted datasets can lead to unreliable LxMs, resulting in vulnerable designs.

\section{Conclusion}
Researchers and practitioners are actively exploring various approaches
to enhance the capabilities of large language models (LLMs) toward handling chip design. 
While LLMs have demonstrated remarkable success \extended{for certain hardware tasks, particularly in HDL and security assertion generation}, other advanced aspects of chip design, like hardware security risks and trust issues, still need to be further explored.
Here, we have reviewed prominent prior art, which are also summarized in our LLM4IC hub.
We further raised several research questions toward important future directions.

\bibliographystyle{IEEEtran}
\bibliography{main}

\end{document}